\documentclass[doublecolumn,letterpaper,11pt,conference]{ieeeconf}

\let\labelindent\relax

\usepackage{breakcites}
\usepackage{amsmath,amssymb,amsfonts}
\usepackage{graphicx}
\usepackage{textcomp}
\usepackage{indentfirst}
\usepackage{rotating}
\usepackage{algorithm,algorithmic}
\usepackage{pgf}
\usepackage{tikz}
\usepackage{tikz-3dplot}
\usepackage{pifont}
\usepackage{aligned-overset}
\usepackage{longtable}
\usepackage{dsfont}
\usepackage{booktabs}
\usepackage{nicefrac}
\usepackage{amsmath}
\usepackage{multirow}

\newcommand{\grad}{\nabla}

\newcommand{\bI}{\mathbf{I}}

\newcommand{\bzero}{\mathbf{0}}

\newcommand{\bx}{\mathbf{x}}

\newcommand{\bz}{\mathbf{z}}

\newcommand{\bepsilon}{{\boldsymbol{\epsilon}}}
\newcommand{\bmu}{{\boldsymbol{\mu}}}

\definecolor{brickred}{rgb}{0.8, 0.25, 0.33}
\definecolor{seagreen}{rgb}{0.18, 0.55, 0.34}
\definecolor{royalblue}{rgb}{0.25, 0.41, 0.88}

\usepackage[usenames,dvipsnames]{pstricks}
\usetikzlibrary{arrows,shapes,automata,decorations.text,decorations.pathmorphing,backgrounds,positioning,fit}
\usepackage{enumitem}
\usepackage{color}
\usepackage{xcolor}
\usepackage{tabularx}
\usepackage{pgfplots}
\usepackage{pst-grad} 
\usepackage{pst-plot} 
\usetikzlibrary{positioning}
\usepackage{rotating}
\usepackage{cite}
\usepackage[colorlinks=true, allcolors=blue]{hyperref}
\usepackage{dblfloatfix}
\usepackage{url,amsmath,setspace,amssymb}
\usepackage{tikz}
\usepackage{graphicx}
\usepackage{caption}
\usepackage{subcaption}
\usepackage{float}
\usepackage{soul}
\usepackage{epsfig}
\usepackage{longtable}
\usepackage{relsize}
\allowdisplaybreaks
\IEEEoverridecommandlockouts
\usepackage{nccmath}

\usepackage{xargs}                      
\usepackage[colorinlistoftodos]{todonotes}

\usetikzlibrary{external}
\definecolor{royalblue}{RGB}{65, 105, 225}
\definecolor{seagreen}{RGB}{46, 139, 87}
\definecolor{firebrick}{RGB}{178,34,34}
\definecolor{darkviolet}{RGB}{138, 43, 226}
\definecolor{carrotorange}{RGB}{237, 145, 33}
\pgfplotsset{every tick label/.append style={font=\small}}

\def\BibTeX{{\rm B\kern-.05em{\sc i\kern-.025em b}\kern-.08em
    T\kern-.1667em\lower.7ex\hbox{E}\kern-.125emX}}
\begin{document}
\title{\LARGE \bf Improving Denoising Diffusion Probabilistic Models via Exploiting Shared Representations}
\author{Delaram Pirhayatifard, Mohammad Taha Toghani,
Guha Balakrishnan,
C\'{e}sar A. Uribe
\thanks{
Department of Electrical and Computer Engineering, Rice University, Houston, TX, USA.
Email addresses: \{\href{mailto:dp43@rice.edu}{dp43}, \href{mailto:mttoghani@rice.edu}{mttoghani},
\href{mailto:guha@rice.edu}{guha}, \href{mailto:cauribe@rice.edu}{cauribe}\}@rice.edu.}}

\allowdisplaybreaks

\maketitle

\begin{abstract}
In this work, we address the challenge of multi-task image generation with limited data for denoising diffusion probabilistic models (DDPM), a class of generative models that produce high-quality images by reversing a noisy diffusion process. We propose a novel method, SR-DDPM, that leverages representation-based techniques from few-shot learning to effectively learn from fewer samples across different tasks. Our method consists of a core meta architecture with shared parameters, i.e., task-specific layers with exclusive parameters. By exploiting the similarity between diverse data distributions, our method can scale to multiple tasks without compromising the image quality. We evaluate our method on standard image datasets and show that it outperforms both unconditional and conditional DDPM in terms of FID and SSIM metrics.
\end{abstract}


\section{Introduction}
\label{sec:introduction}


Diffusion models are a class of generative models that produce high-quality images by reversing a noisy diffusion process \cite{sohl2015deep}. They have shown several advantages over previous state-of-the-art generative models such as GANs \cite{dhariwal2021diffusion}, such as their scalability and their ability to capture the underlying structure of the data, including the spatial relationships between different objects \cite{yang2022diffusion}. This enables them to generate images that are more realistic and diverse than those produced by other generative models \cite{ramesh2022hierarchical,ramesh2021zero}. These advances have made diffusion models powerful and useful tools for generating images and other complex data for various applications, such as computer vision \cite{amit2021segdiff,saharia2022palette}, natural language processing \cite{ramesh2022hierarchical,savinov2021step,li2022diffusion}, artistic image generation \cite{rombach2022text}, medical image reconstruction \cite{peng2022towards}, and music generation \cite{mittal2021symbolic}.


Diffusion models are based on non-equilibrium thermodynamics \cite{sohl2015deep,van2017neural}, where diffusion increases the system’s entropy. They generate samples by gradually introducing random noise to data and learning to reverse the process to obtain the desired data samples. However, diffusion models have some limitations, such as being time-consuming and computationally expensive to train \cite{song2020denoising}, being difficult to troubleshoot and scale to large datasets \cite{chung2022improving,rombach2022high}, and having high-dimensional latent variables similar to the original data. 


Few-shot learning is a type of meta-learning \cite{finn2017model} that enables models to learn from a limited amount of data \cite{vilalta2002perspective,robb2020few}. This is especially useful in scenarios where the data availability is scarce or the training costs and time are high \cite{vilalta2002perspective}. In such cases, few-shot learning can be used to quickly learn from a small number of examples. Several optimization-based and hierarchical-based techniques have been proposed to enable meta-learning for different problems. These techniques allow researchers to make more accurate predictions and to better understand the underlying structure of data. Few-shot learning is also a powerful tool for image generation in limited data setups. It can be used to create a variety of images from a small dataset, such as images of a specific object in different poses or environments. This can be a useful tool for data augmentation, as well as for creating new images for different tasks.


In this work, we study image generation in multi-task setups with limited data per task. We propose to enhance the quality of image generation in diffusion models by leveraging the idea of shared and personalized representations. We introduce a novel hierarchical-based algorithm called \textit{Shared-Representation Denoising Diffusion Probabilistic Model} (SR-DDPM), which exploits a combination of shared and exclusive features \cite{collins2021exploiting} to improve the sample fidelity under limited data regimes. We discuss how our method is capable of fast and light fine-tuning, as well as better scalability to unseen tasks, i.e., data from a new category. We evaluate the performance of SR-DDPM on four standard datasets: MNIST \cite{deng2012mnist}, Fashion-MNIST (FMNIST) \cite{xiao2017fashion}, CIFAR-10 \cite{krizhevsky2009learning}, and CIFAR-100 \cite{krizhevsky2009learning} under limited data samples.


The rest of this paper is structured as follows. Section \ref{sec:background} reviews the related works. Section \ref{sec:alg} introduces the problem setup and our method, \textit{SR-DDPM}, for improving the performance of DDPMs using a mixture of shared and exclusive layers. Section \ref{sec:experiment} presents the numerical results and Section \ref{sec:conclusion} concludes the paper.

\section{Background}
\label{sec:background}


Recent advances in diffusion models have focused on improving the quality and efficiency of the generated images in various ways. For instance, \cite{ho2020denoising,nichol2021improved} introduced the concept of noise-conditioned score networks, which learn the corresponding noise for two consecutive images in the diffusion process. Rombach et al. \cite{rombach2022high} proposed a two-stage approach to distinguish the imperceptible details in high-quality photos via adversarial auto-encoders, which reduce the size of latent DDPM. Moreover, some works proposed non-Markovian and operator learning techniques for implicit fast sampling \cite{song2020denoising,zhang2022gddim,zheng2022fast}.


Ho et al. \cite{ho2022cascaded} found that cascaded diffusion models were capable of generating high-fidelity images without the assistance of auxiliary image classifiers. There have also been recent attempts to boost image quality by incorporating conditional approaches that use noise prediction \cite{giannone2022few,ho2022classifier}. In recent studies, broader corruption processes such as blurring, pixelation, and desaturation have also been considered in training and sampling diffusion models \cite{daras2022soft,bansal2022cold}.


Furthermore, there has been a growing focus on score-based generative modeling using stochastic differential equations (SDE) \cite{vahdat2021score,zeng2022lion,dockhorn2022differentially}, where the goal is to learn score functions, gradients of log probability density functions, on a wide range of noise-perturbed data distributions, and then sample with Langevin-type methods. Additionally, several exceptional efforts have been made for cases with multi-modal datasets and 3D image generation \cite{ramesh2022hierarchical,zeng2022lion,rombach2022high}. This has been achieved by incorporating additional information about the data, such as object labels or scene context, via using attention layers in the model \cite{vaswani2017attention}.

\section{Problem Setup \& Algorithm}
\begin{figure*}
    \centering
    \includegraphics[width=0.55\linewidth]{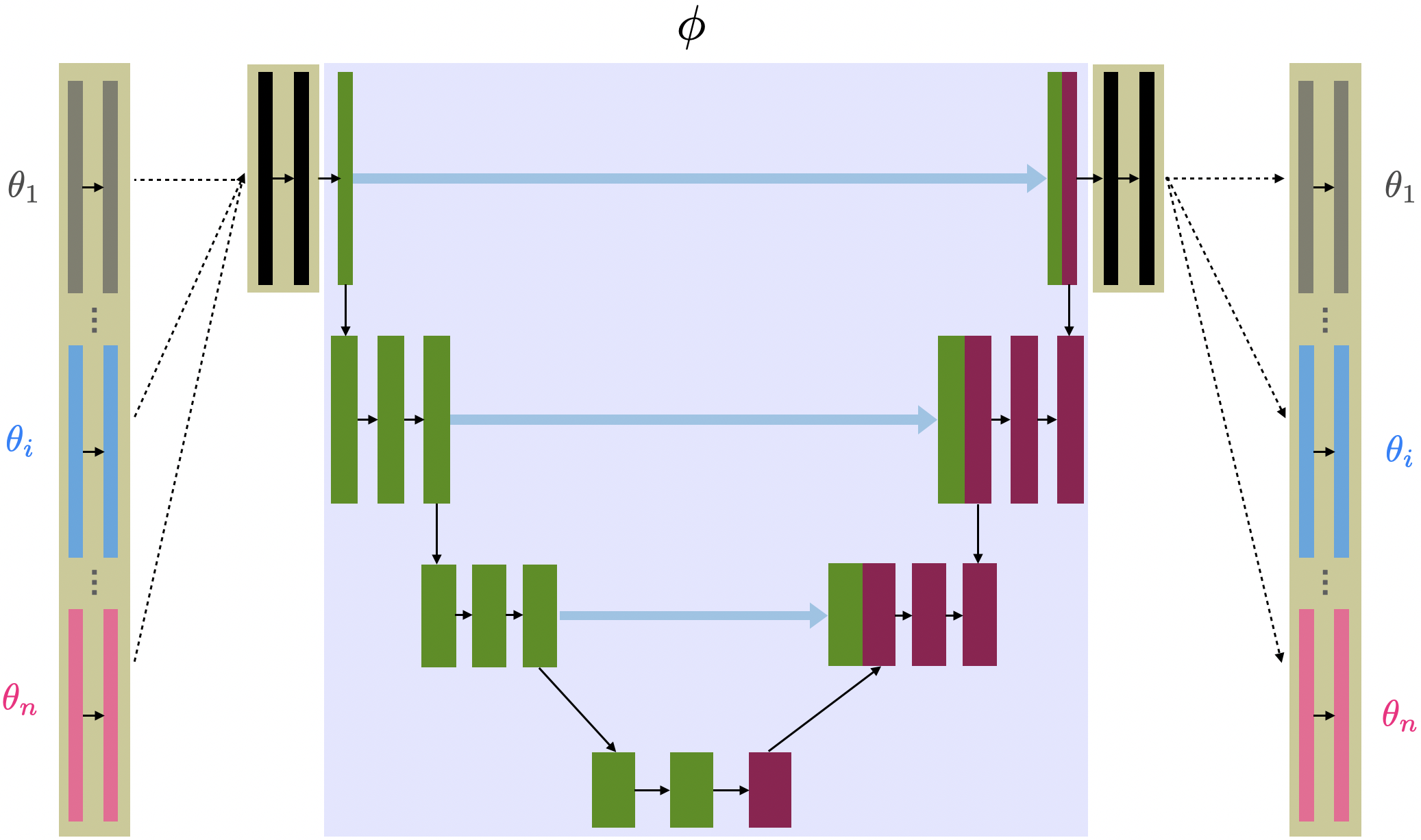}
    \hspace{0.6em}
    \includegraphics[width=0.41\linewidth]{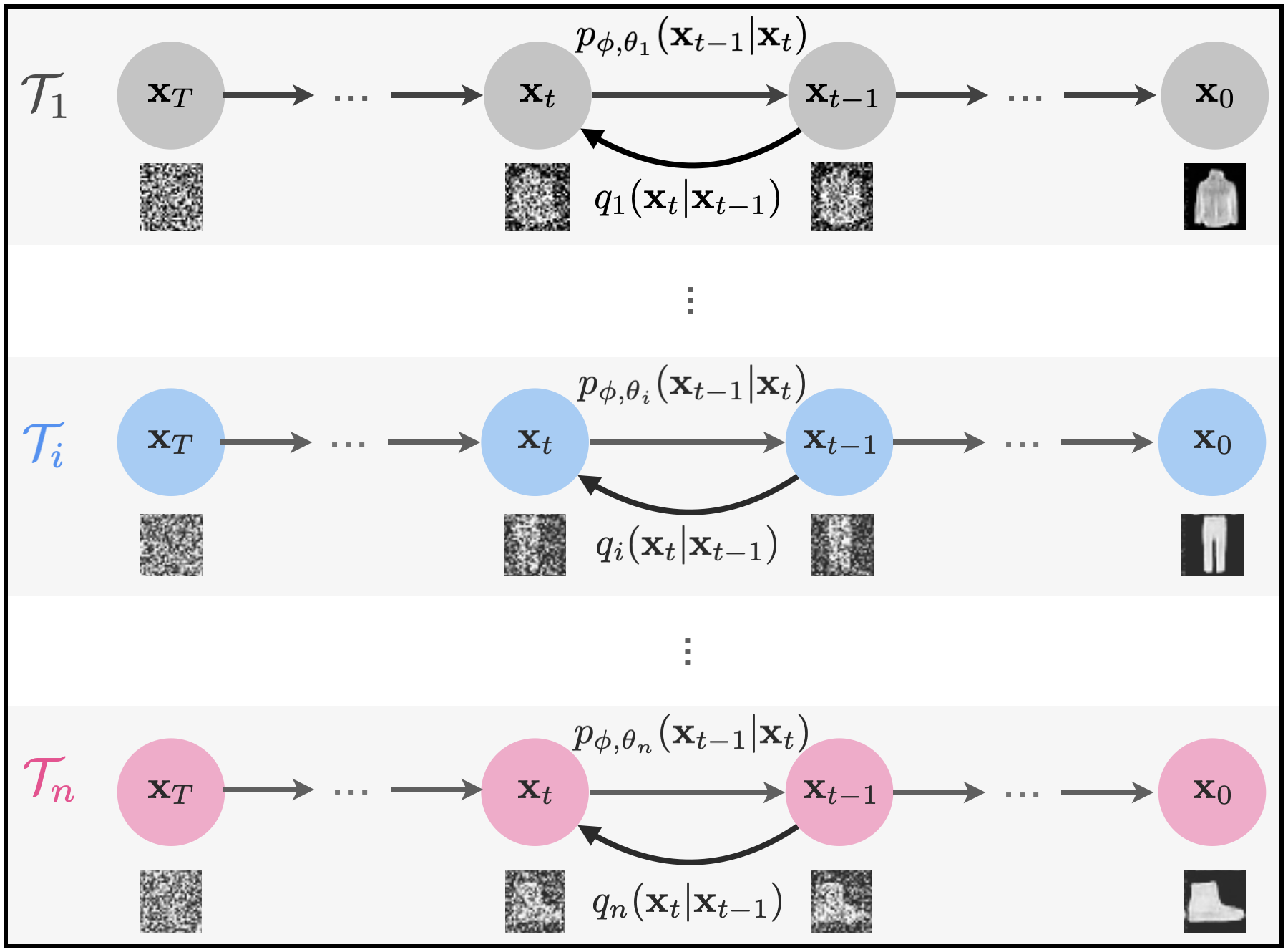}
    \caption{\textbf{(Left)} UNet architecture contains a mixture of shared and exclusive layers. \textbf{(Right)} SR-DDPM for a mixture of $n$ exclusive but similar distributions.}
    \label{fig:unet_fsddpm}
\end{figure*}

\label{sec:alg}
In this section, we first describe the underlying problem setup for few-shot image generation. Then after reviewing the notion and formulation of DDPM \cite{ho2020denoising}, we present our method, SR-DDPM.

\noindent\textbf{Data Setup:} We consider a set of $n$ different tasks \mbox{$\{\mathcal{T}_i\}_{i=1}^n$}, where for each task \mbox{$i\in[n]=\{1,2,\dots,n\}$}, there exist a set of $m_i$ samples \mbox{$\mathcal{S}_i=\{\bx_i^j\}_{j=1}^{m_i}$}, where 
each \mbox{$\bx_i^j{\sim} \mathcal{D}_i$} is an image in a $d$-dimensional space (\mbox{$d=32{\times} 32{\times} 3$} for CIFAR-10 \cite{krizhevsky2009learning}). In the conventional setup for diffusion models, the underlying mechanism is to aggregate all samples $\mathcal{S} = \cup_{i=1}^n \mathcal{S}_i$ irrespective of their task and every $\bx{\in} \mathcal{S}$ is a realization of some generic (global) distribution $\bx{\sim}\mathcal{D}$. \textit{In this research, we unravel how to exploit the combination of diverse yet similar distributions $\mathcal{D}_i$ to improve the quality of image generation in diffusion models.}

We start by stating the problem setup for DDPM and then introduce our method for shared representations.

\noindent\textbf{DDPM:} 
Let $\bx_0\in\mathbb{R}^d$ be an image sampled from distribution $\mathcal{D}$. Moreover, let $\bx_1,\bx_2,\dots,\bx_T$ denote $T$ latent variables where each $\bx_t\in\mathbb{R}^d$, for all $t\in[T]$. The forward (diffusion) process $q$ can be defined as follows:
\begin{align}\label{eq:forward-process}
    q(\bx_{1:T}|\bx_0) &\coloneqq \prod_{t=1}^{T} q(\bx_t|\bx_{t{-}1}),\\
    q(\bx_t|\bx_{t{-}1}) &\coloneqq \mathcal{N}(\bx_t; \sqrt{1{-}\beta_t} \bx_{t{-1}}, \beta_t \bI),
\end{align}
where \mbox{$\beta_1,\beta_2,\dots,\beta_T$} is a variance schedule for the underlying gaussian noise with mean \mbox{$\sqrt{1{-}\beta_t} \bx_{t{-1}}$} and variance \mbox{$\beta_t \bI$} at each timestep $t$. According to \cite{luo2022understanding,ho2020denoising}, the latent variable $\bx_t$ can be directly derived based on the observed data $\bx_0$ as
\begin{align}\label{eq:xt-based-x0}
    \bx_t = \sqrt{\overline{\alpha}_t} \bx_0 + \sqrt{1{-}\overline{\alpha}_t}\bepsilon,
\end{align}
where \mbox{$\overline{\alpha}_t \coloneqq \prod_{s=1}^t \alpha_s$} and \mbox{$\bepsilon{\sim}\mathcal{N}(\bzero,\bI)$}. Moreover, the reverse (generative) process $p_\psi$, parameterized by a set of parameters $\psi$, can be summarized as follows:
\begin{align}\label{eq:reverse-process}
    p_{\psi}(\bx_{0:T}) &\coloneqq \prod_{t=1}^{T} p_{\psi}(\bx_{t{-}1}|\bx_{t}),\\
    p_{\psi}(\bx_{t{-}1}|\bx_{t}) &\coloneqq \mathcal{N}\left(\bx_{t{-}1}; \bmu_{\psi}(\bx_t,t), \sigma_t^2 \bI\right),\\
    \bmu_{\psi}(\bx_t,t)&\coloneqq\frac{1}{\sqrt{\alpha_t}}\left[\bx_t {-} \frac{\beta_t}{\sqrt{1{-}\overline{\alpha}_t}}\bepsilon_{\psi}(\bx_t,t)\right],
\end{align}
where \mbox{$\sigma_t^2 = \beta_t$}, and $\bepsilon_{\psi}: \mathbb{R}^d {\times} \mathbb{N} \to \mathbb{R}^{d}$ is a neural network with parameters $\psi$ that takes $\bx_t$ and timestep $t$ as inputs and estimates the realization of $\bepsilon$ in \eqref{eq:xt-based-x0}. For example, UNet with attention is a proper candidate for $\bepsilon_{\psi}$. In \cite{ho2020denoising}, it is explained that \mbox{$\sigma_t^2 = \tilde{\beta}_t = \frac{1{-}\overline{\alpha}_{t{-}1}}{1{-}\overline{\alpha}_{t}}\beta_t$} provides similar experimental results to \mbox{$\sigma_t^2=\beta_t$}. Note that the underlying assumption in this formulation is that $\bepsilon_{\psi}(\bx_t,t)$ is a shared model across the Markov chain (from $0$ to $T$) which is expressive enough to recover the noise value. Therefore, it is sufficient to optimize the network parameters with respect to some loss function $\mathcal{L}: \mathbb{R}^d {\times} \mathbb{R}^d \to \mathbb{R}^{+}$, i.e., minimizing
\begin{align}\label{eq:loss-generic}
    \mathcal{L}\left(\bepsilon, \bepsilon_{\psi}(\sqrt{\bar\alpha_t} \bx_0 + \sqrt{1{-}\bar\alpha_t}\bepsilon, t) \right),
\end{align}
which quantifies the distance between the original noise and prediction of the denoising model.

Next, we explain SR-DDPM for multi-task denoising diffusion models with shared and exclusive representations.

\begin{algorithm}[b]
\caption{SR-DDPM: Training}\label{alg:training}
\small
\begin{algorithmic}[1]
    \REPEAT
        \STATE{$i\sim \mathrm{Uniform}([n])$} \COMMENT{\textcolor{royalblue}{select a task $\mathcal{T}_i$ from $n$ tasks}}
        \STATE $\bx_0 \sim \mathcal{D}_i$ \COMMENT{\textcolor{royalblue}{sample data $\bx_0$ from task $\mathcal{T}_i$}}
        \STATE $t \sim \mathrm{Uniform}([T])$ \COMMENT{\textcolor{royalblue}{sample timestep $t$}}
        \STATE $\bepsilon\sim\mathcal{N}(\bzero,\bI)$
        \STATE Compute the gradient and apply one step of optimizer:
        $$\qquad \grad_{\phi,\theta_i} \mathcal{L}\left(\bepsilon, \bepsilon_{\phi,\theta_i}(\sqrt{\bar\alpha_t} \bx_0 + \sqrt{1{-}\bar\alpha_t}\bepsilon, t) \right)$$
    \UNTIL{converged}
\end{algorithmic}
\end{algorithm}

\begin{algorithm}[b]
\caption{SR-DDPM: Sampling} \label{alg:sampling}
\small
\begin{algorithmic}[1]
    \vspace{0.4em}
    \STATE{$i\sim \mathrm{Uniform}([n])$} \COMMENT{\textcolor{royalblue}{select a task $\mathcal{T}_i$ from $n$ tasks}}
    \STATE $\bx_T \sim \mathcal{N}(\bzero, \bI)$ \COMMENT{\textcolor{royalblue}{sample a noise signal $\bx_T$}}
    \FOR{$t=T, \dotsc, 1$}
        \STATE{$\bz \sim \mathcal{N}(\bzero, \bI)$ if $t > 1$, else $\bz = \bzero$}
        \STATE{$\bx_{t-1} = \frac{1}{\sqrt{\alpha_t}}\left( \bx_t - \frac{1-\alpha_t}{\sqrt{1-\bar\alpha_t}} \bepsilon_{\phi,\theta_i}(\bx_t, t) \right) + \sigma_t \bz$}\label{ln:samp-denoise}
    \ENDFOR
    \STATE \textbf{return} $\bx_0$
    \vspace{.04in}
\end{algorithmic}
\end{algorithm}

\noindent\textbf{SR-DDPM:}
Our goal is to exploit the exclusiveness of each task $\{\mathcal{T}_i\}_{i=1}^n$ by splitting the denoising network architecture into shared and personal (exclusive) layers. Figure \ref{fig:unet_fsddpm} depicts the UNet structure that uses a common set of parameters $\phi$ for all tasks $i{\in}[n]$, and a distinct set of parameters $\{\theta_i\}_{i=1}^n$ for each task. The set of parameters $\psi$ in \eqref{eq:reverse-process} is the combination of $\phi$ and $\theta_i$, for any $i\in[n]$. This allows us to jointly capture both shared and unshared features. For example, Figure \ref{fig:unet_fsddpm} shows that for different tasks involving various outfits, we train and sample from $n$ parallel Markov chains with shared parameters $\phi$ and exclusive parameters $\{\theta_i\}_{i=1}^n$. In other words, we minimize the following:
\begin{align}\label{eq:loss-fs}
\mathbb{E}_{i}\left[\mathcal{L}\left(\bepsilon, \bepsilon_{\phi,\theta_i}(\sqrt{\bar\alpha_t} \bx_0^i + \sqrt{1{-}\bar\alpha_t}\bepsilon, t) \right)\right],
\end{align}
where $\bx_0^i\sim\mathcal{D}_i$ and $i\sim\mathrm{Uniform}([n])$. Algorithms \ref{alg:training} and \ref{alg:sampling} respectively describe the training and sampling processes of SR-DDPM. As shown in Algorithm \ref{alg:training}, at  the training phase, we randomly choose a task and an image from that task. Then, we use a first-order optimization method such as Adam \cite{kingma2014adam} to optimize the stochastic gradient and minimize the cost in \eqref{eq:loss-fs}. Finally, we generate samples by feeding a noise signal to the network and applying the denoising process of Algorithm \ref{alg:sampling}.


\section{Experiments}
\label{sec:experiment}



In this section, we describe the experimental setup and the results of our proposed method. We compare our method with unconditional and conditional DDPM. 

We consider four standard datasets: MNIST, FMNIST, CIFAR-10, and CIFAR-100. We implement a multi-task scheme with $500$ samples per task. Following Section \ref{sec:alg}, we adopt a UNet as the denoising network with four layers for all methods. The network has a bottleneck in the middle to learn only the most important features of the data. We increase the number of channels by a factor of two and decrease the image size by the same factor per layer. We personalize one layer as the exclusive stage at the first and end of the network for all datasets. We train the model for each method within $600$ epochs. We use the Adam optimizer with a learning rate of $5{\times}10^{-4}$ for all experiments.

\begin{table}[t]
\begin{center}
\captionof{table}{Comparison of the performance of Denoising Diffusion Probabilistic Models (DDPM), Conditional DDPM (C-DDPM), and Shared-Representation DDPM (SR-DDPM) for \mbox{$T{=}500$}, \mbox{$4$-layer} UNet with one exclusive layer, linear $\beta$ schedule, and $600$ training epochs.}
\label{tab:result}
\begin{tabular}{lclcc}
    \toprule
    Dataset & $|\{\mathcal{T}_i\}_i|$ & Method & FID $\downarrow$ & SSIM $\uparrow$\\
    \midrule
    \multirow{4}{*}{MNIST} &\multirow{4}{*}{$10$} & DDPM& 3.67  & 0.881 \\
    \cmidrule(lr){3-5}
    && C-DDPM & 2.14 & 0.884 \\
    \cmidrule(lr){3-5}
    && SR-DDPM & \textbf{2.04} & \textbf{0.887} \\
    \hline
    \multirow{4}{*}{FMNIST} &\multirow{4}{*}{$10$} & DDPM& 4.80 & 0.908 \\
    \cmidrule(lr){3-5}
    && C-DDPM & 2.72 & \textbf{0.915} \\
    \cmidrule(lr){3-5}
    && SR-DDPM & \textbf{2.48} & 0.909 \\
    \midrule
    \multirow{4}{*}{CIFAR-10} & \multirow{4}{*}{$10$} & DDPM& 12.64 & 0.946 \\
    \cmidrule(lr){3-5}
    && C-DDPM & 12.86  & \textbf{0.949} \\
    \cmidrule(lr){3-5}
    && SR-DDPM & \textbf{10.87} & \textbf{0.949} \\
    \midrule
    \multirow{4}{*}{CIFAR-100} &\multirow{4}{*}{$20$} & DDPM& 13.74 & \textbf{0.944} \\
    \cmidrule(lr){3-5}
    && C-DDPM & 11.54  & 0.942 \\
    \cmidrule(lr){3-5}
    && SR-DDPM & \textbf{11.30} & \textbf{0.944} \\
    \bottomrule
    
\end{tabular}
\end{center}
\end{table}



We quantitatively compare the performance of our model with DDPM and Conditional DDPM (C-DDPM) using the implementation of DDPM \cite{ho2020denoising} on Hugging Face \cite{rogge2022annotated}. We measure the performance of SR-DDPM on the four different datasets using sample quality (FID@$10$k) and structural similarity (SSIM) on test data. Table \ref{tab:result} compares SR-DDPM with unconditional and conditional DDPM. Our method achieves better FID scores than the other two on all four datasets.

\begin{figure}[t]
    \centering
    \begin{minipage}{0.48\linewidth}
    \includegraphics[width=1\linewidth]{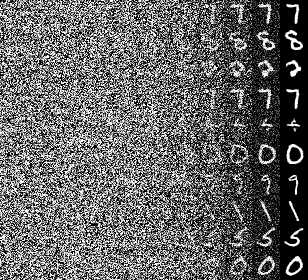}
    \centering (a) MNIST
    \end{minipage}
    \begin{minipage}{0.48\linewidth}
    \includegraphics[width=1\linewidth]{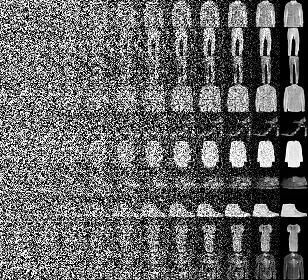}
    \centering (b) FMNIST
    \end{minipage}
    \begin{minipage}{0.48\linewidth}
    \includegraphics[width=1\linewidth]{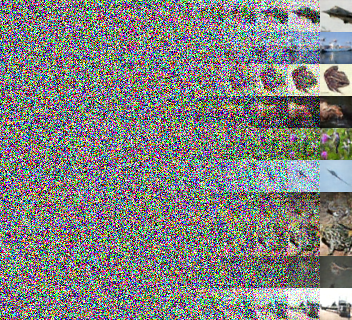}
    \centering (c) CIFAR-10
    \end{minipage}
    \begin{minipage}{0.48\linewidth}
    \includegraphics[width=1\linewidth]{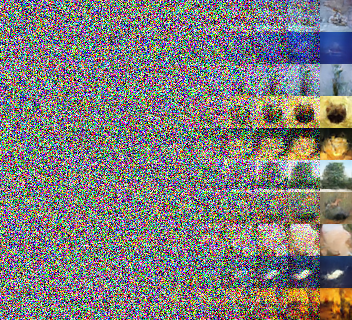}
    \centering (d) CIFAR-100
    \end{minipage}
    \caption{The process of diffusion image reconstruction using SR-DDPM. For each dataset, we generate $10$ samples with $T=500$ and visualize the reconstructed image at \mbox{$t = 0, 50, 100, \dots, 500$}.}
    \label{fig:diffusion-generation}
\end{figure}

\begin{figure*}
    \centering
    \includegraphics[width=\linewidth]{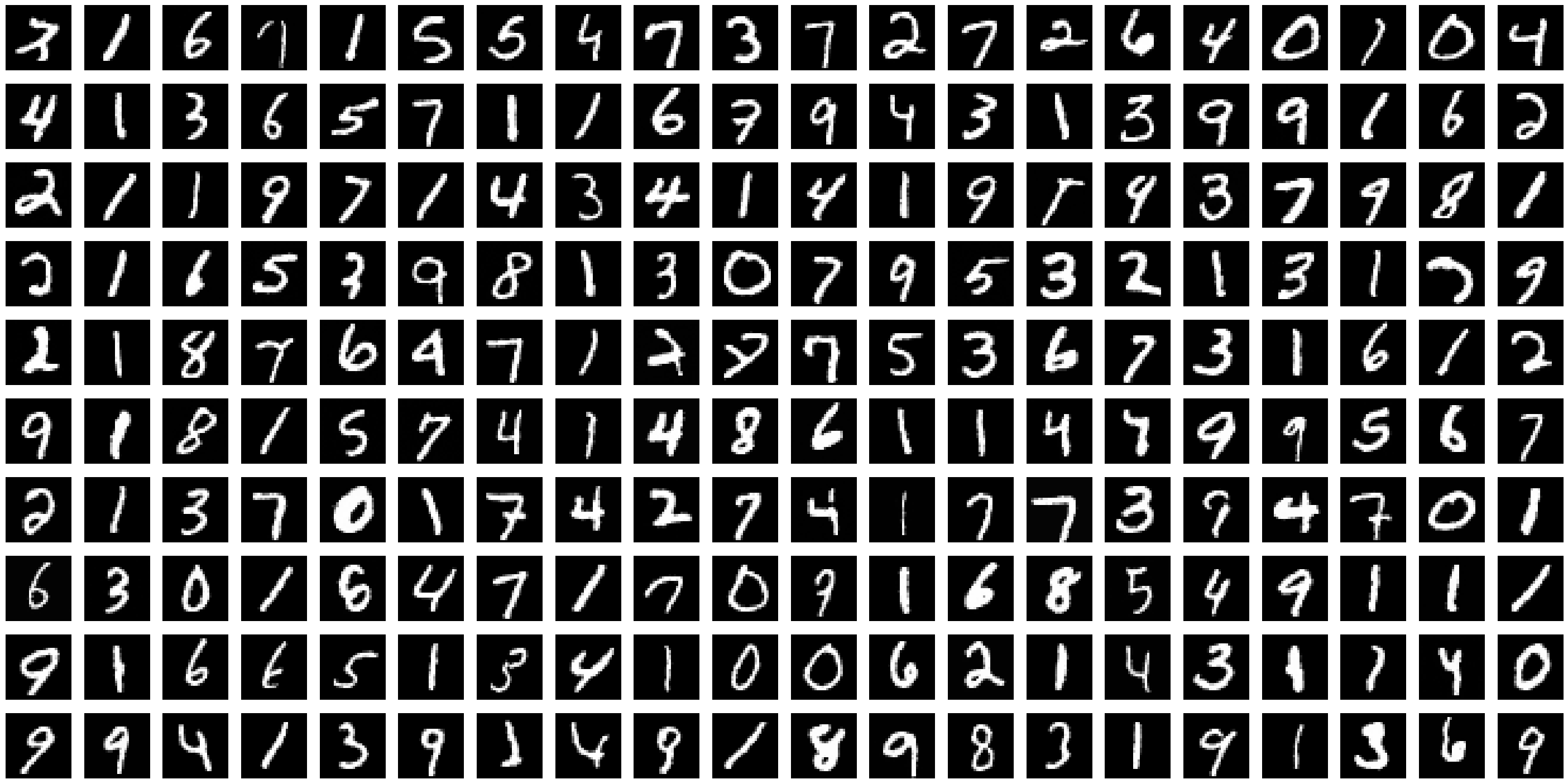}
    \caption{Samples generated via \textbf{DDPM} for \textbf{MNIST}.}
    \label{fig:mnist-DDPM}
\end{figure*}

\begin{figure*}
    \centering
    \includegraphics[width=\linewidth]{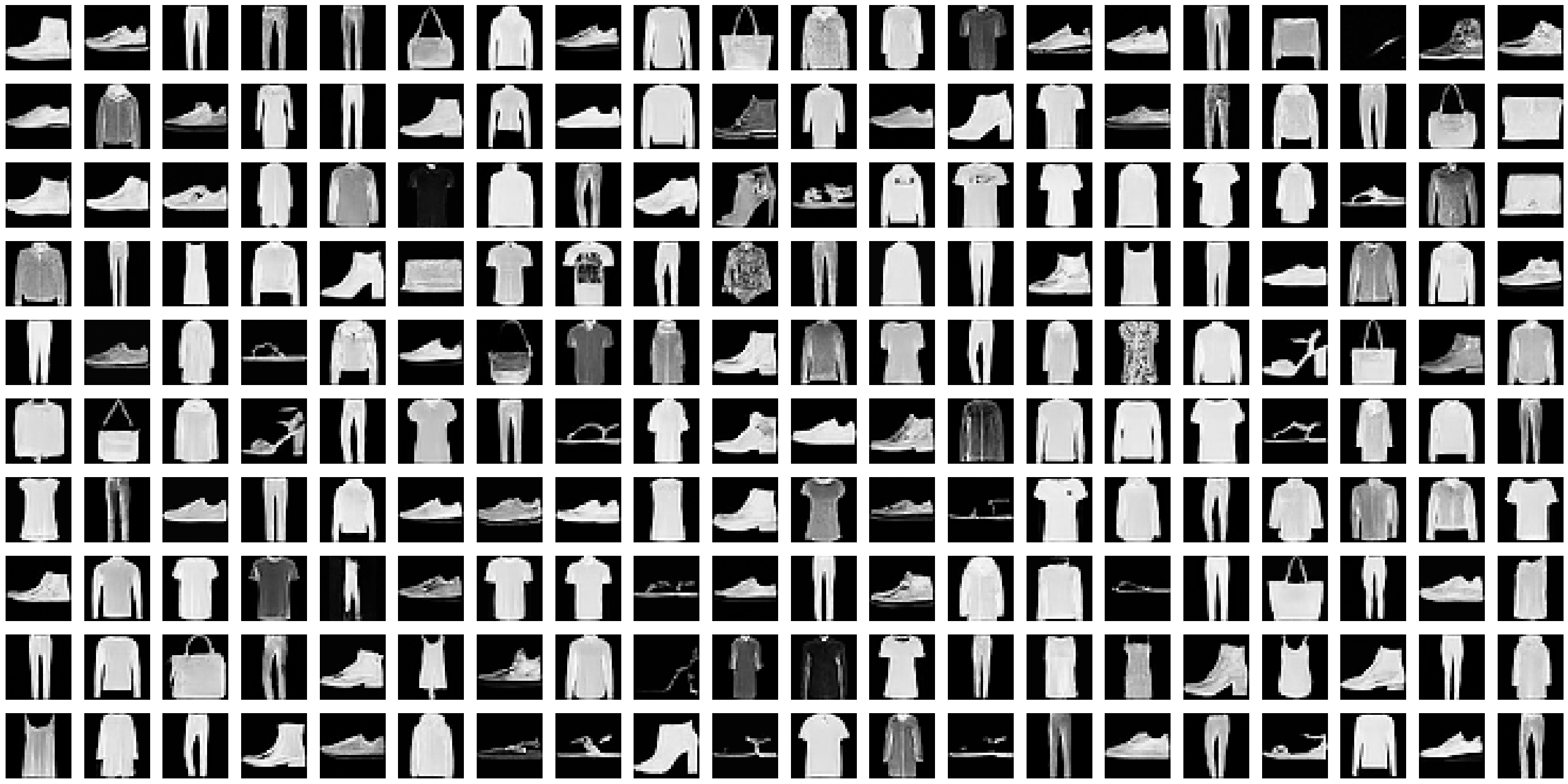}
    \caption{Samples generated via \textbf{DDPM} for \textbf{FMNIST}.}
    \label{fig:fmnist-DDPM}
\end{figure*}

\begin{figure*}
    \centering
    \includegraphics[width=\linewidth]{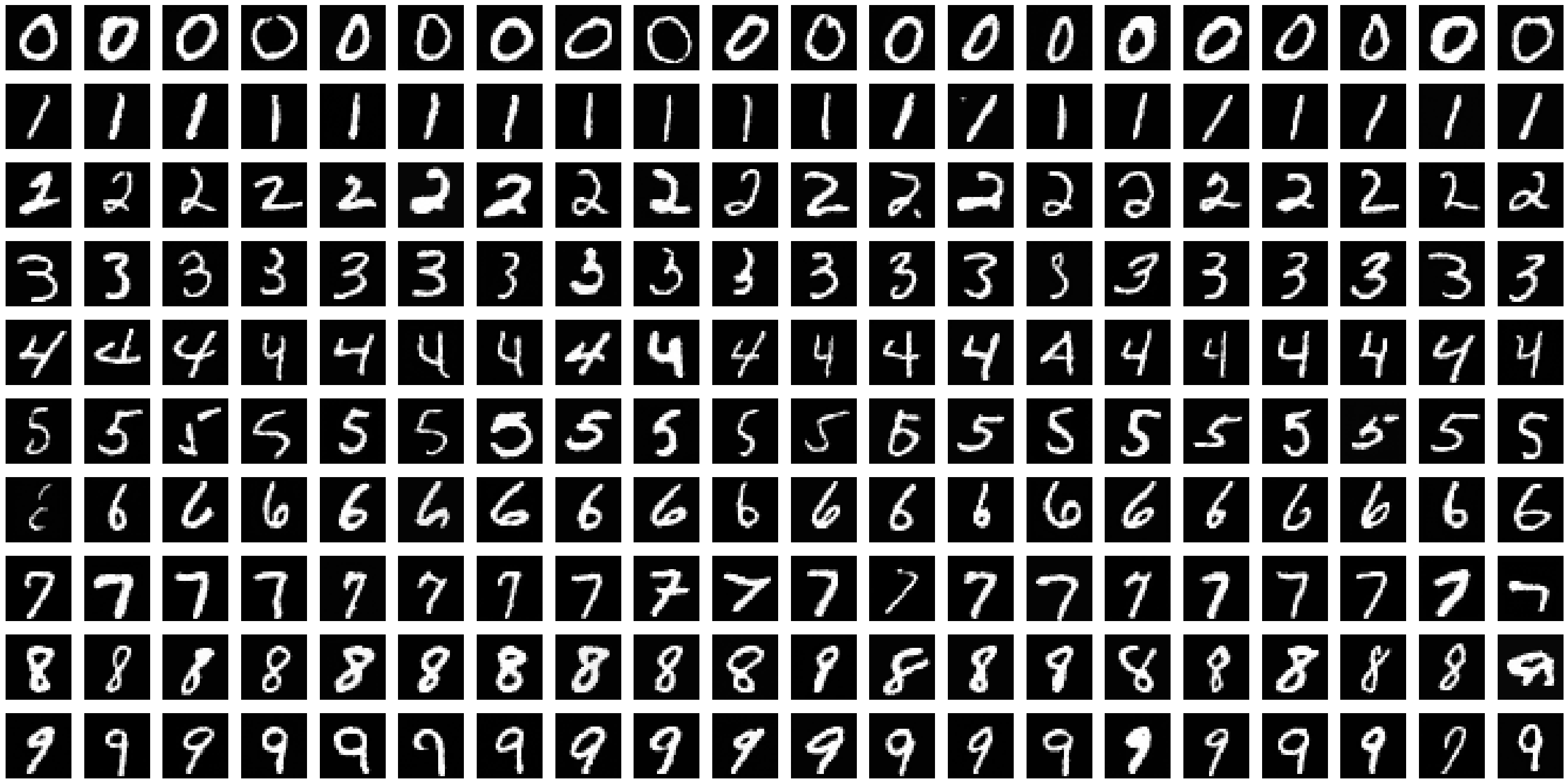}
    \caption{Each row shows 20 samples generated via \textbf{SR-DDPM} through the mixture of exclusive layers $\{\theta_i\}_{i\in[n]}$ and shared layer $\phi$ for \textbf{MNIST}.}
    \label{fig:mnist-SR-DDPM}
\end{figure*}

\begin{figure*}
    \centering
    \includegraphics[width=\linewidth]{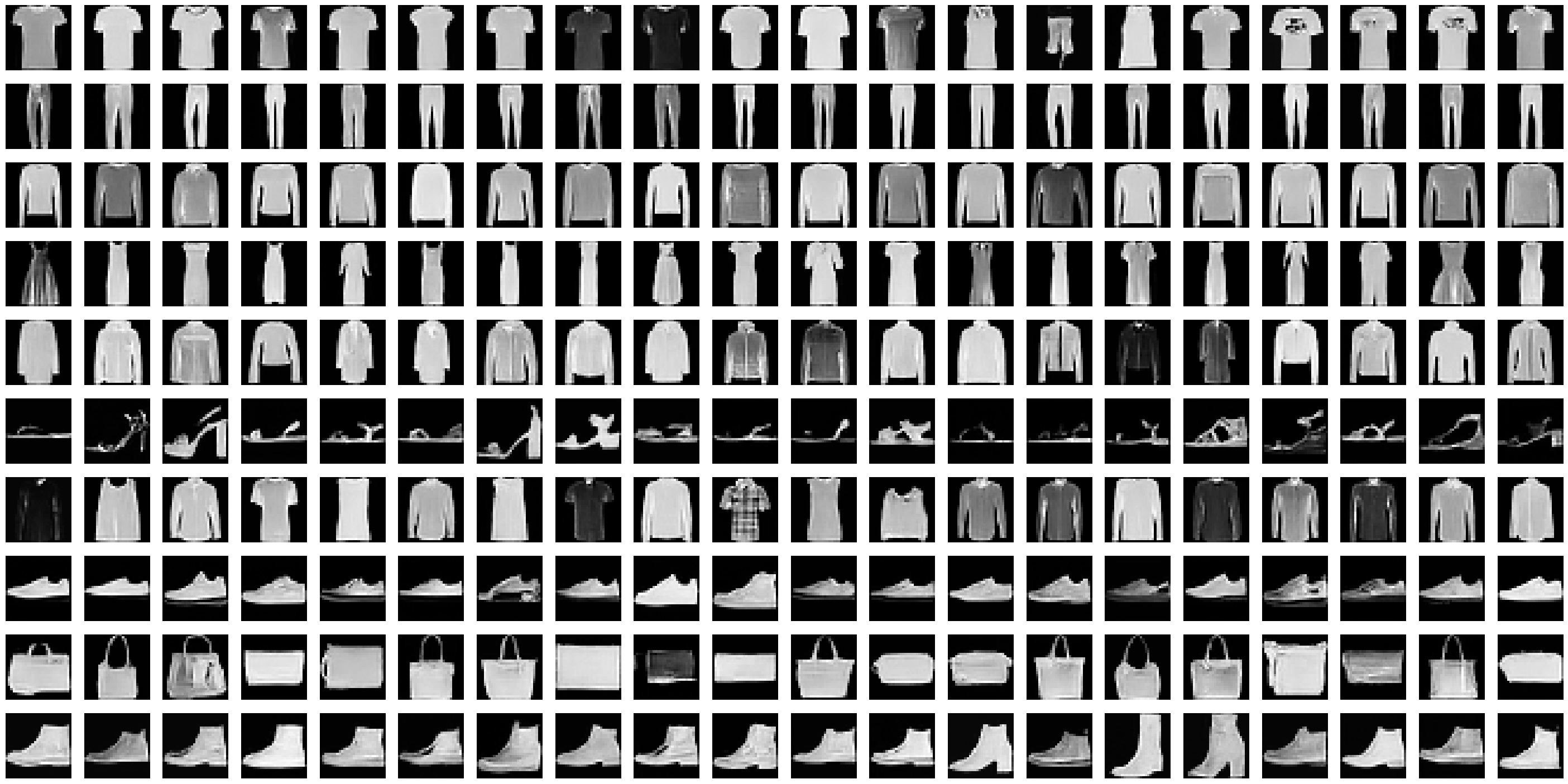}
    \caption{Each row shows 20 samples generated via \textbf{SR-DDPM} through the mixture of exclusive layers $\{\theta_i\}_{i\in[n]}$ and shared layer $\phi$ for \textbf{FMNIST}.}
    \label{fig:fmnist-SR-DDPM}
\end{figure*}

Figure \ref{fig:diffusion-generation} visualizes the reverse process for image generation for \mbox{$T=500$} on all datasets. We also display $20$ samples from each task in Figures \ref{fig:mnist-DDPM}-\ref{fig:fmnist-SR-DDPM}. The images generated from FMNIST show that the trained model can identify similarities between the tasks. For example, look at the T-shirt and Dress generation in Figure \ref{fig:fmnist-SR-DDPM}. Some of the images generated from one task overlap with the other task when using the corresponding exclusive layers. This implies that the method can capture the similarity between the tasks implicitly by using the shared and exclusive layers.

\section{Conclusion}
\label{sec:conclusion}
We presented a novel algorithm for training diffusion models with limited data. Our method outperforms unconditional and conditional DDPM on image generation in the same training time. We also found that the personal layer for each task can detect similarities among tasks automatically. This means that we can train a new personal layer for a new task without fine-tuning the whole network. Our method offers an interpretable way to generate images by using a combination of shared and unshared parameters to capture the differences among tasks.


\bibliographystyle{IEEEtran}
\bibliography{ref}

\clearpage

\end{document}